%% file: main.tex
\documentclass[]{style}

\usepackage[toc,page,header]{appendix}
\usepackage{enumitem}

\input{resources/packages}
\title{HumanNet: Scaling Human-centric Video Learning to One Million Hours}

\author{
  Yufan Deng\texorpdfstring{$^{}$}{} \hspace{0.1cm}
  Daquan Zhou\texorpdfstring{$^{}$}{}
}

\makeatletter
\g@addto@macro\authorlist{\\[3mm]
  {\authorfont\sffamily DAGroup\texorpdfstring{$^{}$}{} \hspace{0.1cm} SimpleSilicon Innovation Team\texorpdfstring{$^{}$}{}}%
}
\makeatother

\affiliation[]{Peking University}
\newcommand{\answerTODO}[1][]{\textcolor{red}{\bfseries [TODO]}}
\newcommand{\justificationTODO}[1][]{\textcolor{red}{\bfseries [TODO]}}

\input{sec/0_abstract}

\checkdata[
\raisebox{-0.2em}{\includegraphics[width=0.025\linewidth]{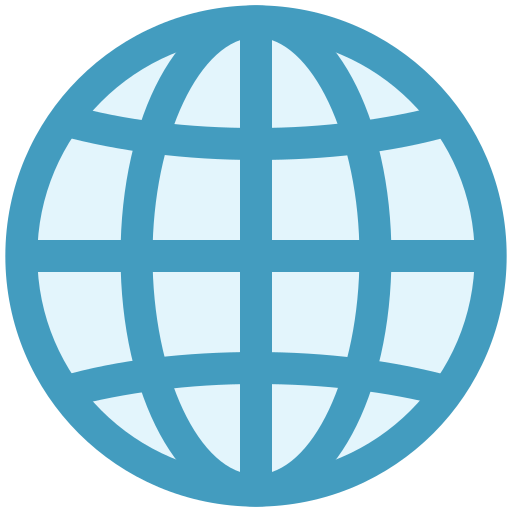}}~~Project Page]{\href{https://dagroup-pku.github.io/HumanNet/}{\texttt{https://dagroup-pku.github.io/HumanNet/}}
\\[-1.5ex]}

\checkdata[
\raisebox{-0.2em}{\includegraphics[width=0.025\linewidth]{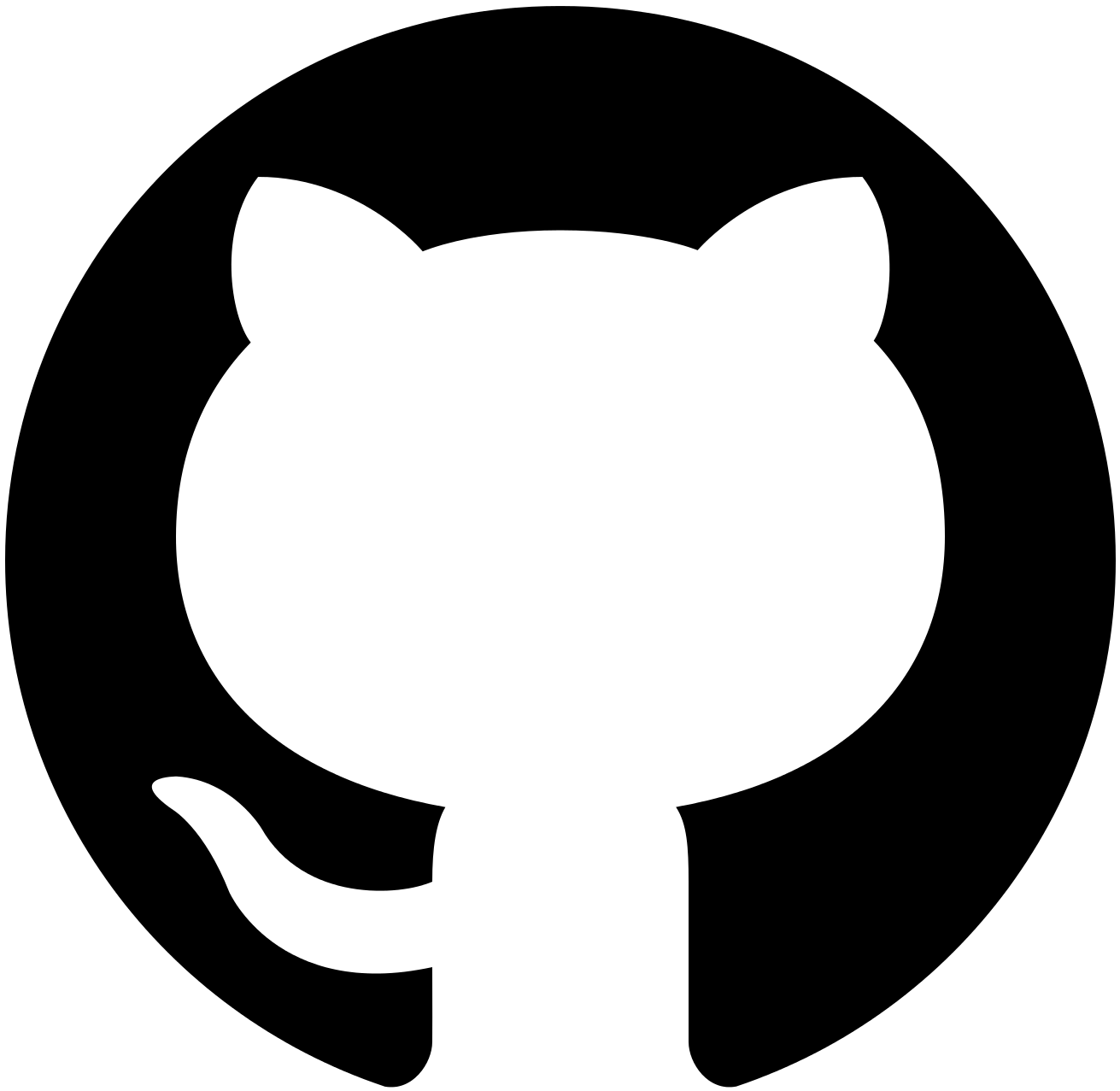}}~~GitHub Repo]{\href{https://github.com/DAGroup-PKU/HumanNet/}{\texttt{https://github.com/DAGroup-PKU/HumanNet/}}
\\[-1.7ex]}

\checkdata[
\raisebox{-0.2em}{\includegraphics[width=0.025\linewidth]{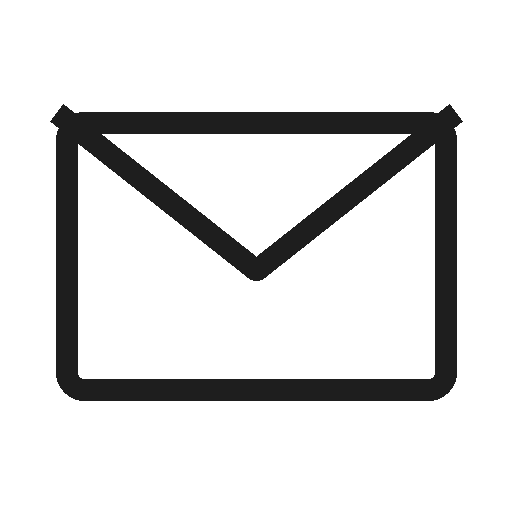}}~~Corresponding Author]{\href{mailto:zhoudaquan21@gmail.com}{\texttt{Daquan Zhou}}
\\[-1.7ex]}

\begin{document}
\maketitle
% \begingroup\let\thefootnote\relax\footnotetext{For any question, please contact Yufan Deng at \href{mailto:dengyufan10@stu.pku.edu.cn}{\texttt{dengyufan10@stu.pku.edu.cn}}.}\endgroup

\begin{figure}[ht]
\centering
\includegraphics[width=\linewidth]{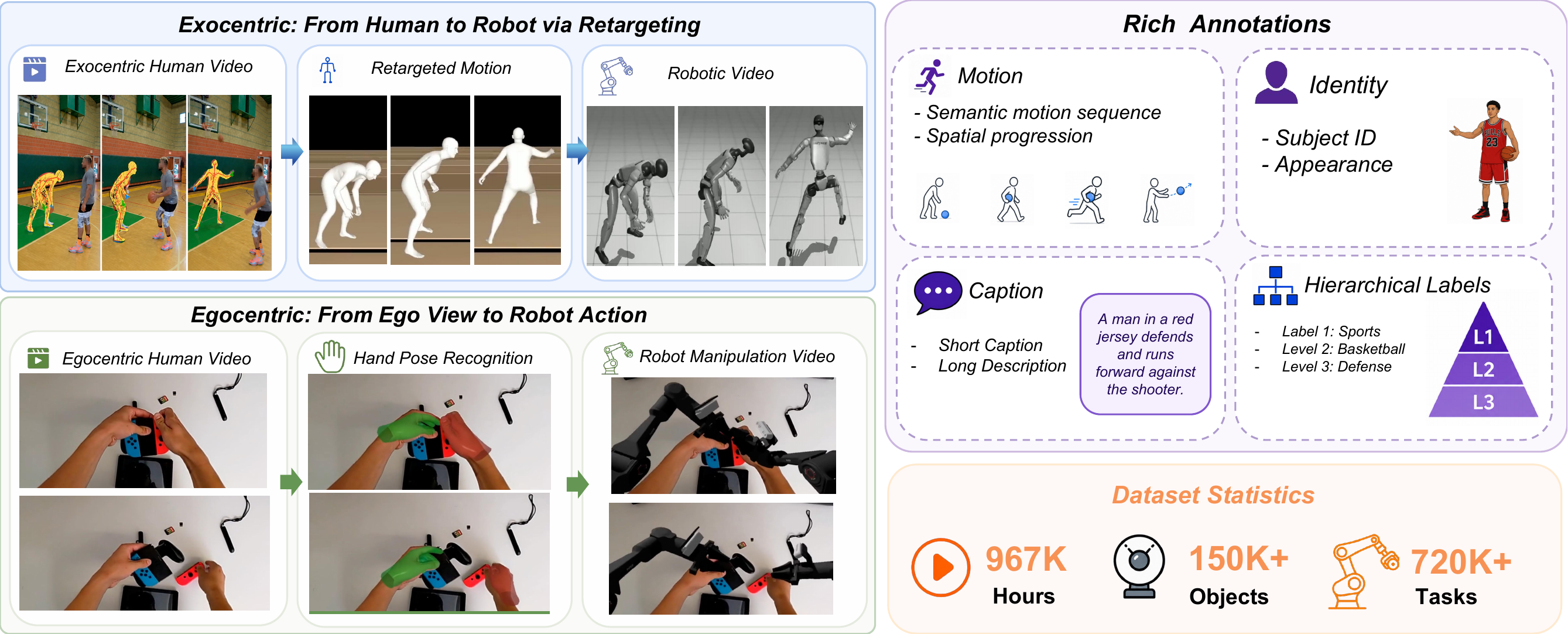}
\caption{Overview of \textcolor{pkured}{HumanNet}, a one-million-hour human-centric video corpus for embodied learning. \textbf{Left:} two viewpoint-specific bridges from human video to robot supervision, where exocentric video is converted into robot motion through retargeting, while egocentric video is paired with hand pose for manipulation transfer. \textbf{Right:} each clip is enriched with motion, identity, caption, and hierarchical-label annotations, and the corpus is summarized by headline statistics on duration, object diversity, and task coverage.}
\label{fig:teaser}
\end{figure}

\input{sec/1_intro}
\input{sec/2_related_work}
\input{sec/3_dataset}
\input{sec/4_downstream}
\input{sec/5_conclusion}
\input{sec/6_limitation}

\clearpage

\bibliographystyle{plainnat}
\setlength{\bibhang}{0pt}
\setlength\bibindent{0pt}
\bibliography{main}

% \clearpage

% \beginappendix
% \input{sec/X_suppl}

\end{document}

%% file: resources/packages.tex
\usepackage{natbib}

\usepackage{CJKutf8}

\usepackage{xargs}  

\usepackage{todonotes}  

\usepackage{multirow}

\usepackage{cleveref}

\usepackage{amsmath}
\usepackage{dsfont}

\usepackage{svg}

\usepackage{mathrsfs}
\usepackage{adjustbox}
\usepackage{multirow}
\usepackage{multicol}
\usepackage{tcolorbox}
\usepackage{changepage}
\usepackage{enumitem}
\usepackage{graphicx}
\usepackage{amssymb}
\usepackage{xcolor}
\usepackage{float}
\usepackage{multirow}
\usepackage{threeparttable}
\usepackage{graphicx}
\usepackage{subcaption}
\usepackage{algorithm}
\usepackage{algpseudocode}
\usepackage{wrapfig}
\usepackage[table]{xcolor}  % loads xcolor with table support
\usepackage{colortbl}       % gives \columncolor, \rowcolor
\usepackage{tabularx} % Add to your preamble
\usepackage{makecell} % Add to your preamble
\usepackage[dvipsnames]{xcolor}

\newcolumntype{g}{>{\columncolor{gray!10}}c} % gray background

\definecolor{catgray}{gray}{0.9}
\definecolor{skyblue}{rgb}{0.53,0.81,0.92} % sky blue

\colorlet{skyblue!30}{skyblue!30!white} % 30% skyblue, 70% white

\definecolor{customblue}{RGB}{70,130,180}  % This is equivalent to rgb(70,130,180)

\newtcolorbox{evolbox}[2][]{%
  enhanced,
  colframe=customblue,
  colback=white,
  coltitle=white,
  rounded corners,
  boxrule=1pt,
  titlerule=0pt,
  toptitle=1mm,
  bottomtitle=1mm,
  fonttitle=\bfseries,
  width=#2\textwidth, % This takes the second parameter as the width fraction
  #1
}
\usepackage{minitoc}
\usepackage{url}

\PassOptionsToPackage{table,xcdraw}{xcolor}
\usepackage{titletoc}
\usepackage{placeins}
\usepackage{pifont}

\definecolor{RowBlue}{HTML}{E9F2FB}
\definecolor{RowRed}{HTML}{F9EAEA}
\definecolor{Top1}{HTML}{50DB4B} % 深绿
\definecolor{Top2}{HTML}{A5FFA2} % 中绿
\definecolor{Top3}{HTML}{D9FFD9} % 浅绿
\definecolor{Sub1}{HTML}{EAB8B8}
\definecolor{Sub2}{HTML}{E4E4E4}

\renewcommand{\emph}[1]{\textit{#1}}

%% file: sec/0_abstract.tex
\abstract{
Progress in embodied intelligence increasingly depends on scalable data infrastructure. While vision and language have scaled with internet corpora, learning physical interaction remains constrained by the lack of large, diverse, and richly annotated human activity data.
We present \textbf{\textcolor{pkured}{HumanNet}}, a one-million-hour human-centric video corpus that captures how humans interact with the physical world at scale. HumanNet spans both first-person and third-person perspectives and covers fine-grained activities, human-object interactions, tool use, and long-horizon behaviors across diverse real-world environments. Beyond raw video, the dataset provides interaction-centric annotations, including captions, motion descriptions, and hand and body-related signals, enabling motion-aware and interaction-aware learning.
Beyond scale, HumanNet introduces a systematic data curation paradigm for embodied learning, where human-centric filtering, temporal structuring, viewpoint diversity, and annotation enrichment are treated as first-class design principles. This design transforms unstructured internet video into a scalable substrate for representation learning, activity understanding, motion generation, and human-to-robot transfer.
We conduct a first-step validation on the value of this design through controlled vision-language-action ablation: under a fixed set of validation data, continued training from the Qwen VLM model with 1{,}000 hours of egocentric video drawn from HumanNet surpasses the continued training with 100 hours of real-robot data from Magic Cobot, indicating that egocentric human video could
be a scalable and cost-effective substitute for robot data.
By building this project, we aim to explore the opportunity to scale embodied foundation models using human-centric videos, rather than relying solely on robot-specific data.
}

%% file: sec/1_intro.tex
\section{Introduction}\label{sec:intro}
Embodied learning systems are still data-limited. In language and vision-language modeling, recent foundation models continue to improve by scaling model capacity together with massive, heterogeneous text, image, and multimodal web data~\cite{deepseekv3,qwen3,qwen25vl,internvl25,gemma3,phi4multimodal}. By contrast, physical interaction models are still typically trained on collections that are orders of magnitude smaller, narrowly focused on a handful of benchmark tasks, and often tied to a specific robot platform, control interface, or sensing stack~\cite{openx,droid,rt1,rt2}. This mismatch in scale has become one of the clearest bottlenecks for general-purpose embodied intelligence.

Human-centric video offers a promising alternative, as large-scale human activity and instructional video corpora have long served as a foundation for visual representation learning, temporal reasoning, and action understanding~\cite{activitynet,kinetics,charades,ava,something,howto100m}. Humans naturally perform rich manipulation, tool use, locomotion, navigation, social coordination, and multi-step procedural activities across homes, workplaces, shops, kitchens, warehouses, public spaces, and outdoor settings. First-person video preserves the viewpoint from which actions are executed, exposing contact dynamics, hand-object relations, temporal intent, and the visual consequences of motor decisions. Third-person video complements this signal by making full-body motion, posture, interaction context, surrounding agents, and scene-level dynamics easier to observe. Large-scale community resources such as Ego4D~\cite{ego4d}, EPIC-KITCHENS~\cite{egokitchens}, Ego-Exo4D~\cite{egoexo4d}, and EgoSchema~\cite{egoschema} have expanded recognition, forecasting, narration, and multimodal understanding from egocentric and paired exocentric video, while structured interaction resources such as HOI4D~\cite{hoi4d} show the value of dense hand-object supervision. Recent work has shown that human-centered data can improve robot learning and representation learning~\cite{r3m,egomimic,egoscale,egoverse,deng2026rethinking}, but current corpora remain limited in duration, fragmented across collection efforts, or optimized for a narrow set of downstream tasks.

Our framing is informed by recent dataset and robot-learning efforts. EgoScale~\cite{egoscale} demonstrates that scaling egocentric human data can produce predictable gains for dexterous manipulation, while EgoVerse~\cite{egoverse} shows the value of a shared ecosystem for continuously growing egocentric robot-learning data across institutions. Ego-Exo4D~\cite{egoexo4d} further motivates pairing first-person and third-person views to recover both actor-centered intent and scene-centered physical context. The Being-H line of work~\cite{beingh0,beingh05,beingh07} argues that human interaction traces can function as a scalable substrate for cross-embodiment learning when coupled with unified representations. Complementary systems co-train imitation policies on aligned human egocentric traces and robot demonstrations~\cite{egomimic}, and open vision-language-action stacks increasingly mix heterogeneous robot logs with human video at foundation-model scale~\cite{gr00t}, alongside large scripted multi-skill robot corpora~\cite{rh20t}. 
\textit{Building on this perspective, we focus on the dataset itself: how to define scope beyond a single viewpoint, structure a taxonomy, curate sources, characterize scale, and articulate the downstream value of a corpus that is large enough to matter for physical AI.}

This paper advocates a data-centric answer to that limitation: scale human-centric video aggressively, while treating curation, viewpoint diversity, and annotation taxonomy as core scientific contributions rather than bookkeeping. We introduce a one-million-hour corpus of human-centric video and describe the design choices required to turn heterogeneous first-person and third-person footage into a pretraining-ready resource, as illustrated in Figure~\ref{fig:teaser}. As the largest human video dataset to date, it is not merely large; rather, it is designed to provide breadth over activities, environments, objects, body motions, interaction styles, and camera viewpoints while preserving enough physical structure to support fine-grained human activity understanding, motion-aware representation learning, procedural reasoning, and human-to-robot transfer.

To verify that this design translates into measurable downstream value, we further conduct a controlled validation under a unified vision-language-action post-training protocol. Holding the policy architecture and the downstream corpus fixed, we vary only the pretraining source, and find that 1{,}000 hours of egocentric video drawn from \emph{HumanNet} attains validation loss on par with, and on several task groups below, that of a model initialized from 100 hours of real-robot data. \textbf{This result substantiates the central claim of \emph{HumanNet}: large-scale egocentric human video is not merely a complementary visual corpus, but a scalable and cost-effective substitute that narrows the gap between internet-scale perception and embodied action learning.}

Table~\ref{tab:comparison} provides an illustrative side-by-side view of \emph{HumanNet} against representative prior corpora along dimensions that matter for human-centric video learning and embodied pretraining: duration, viewpoint coverage, activity scope, and the intended path to embodied use. The comparison is intended to communicate the order-of-magnitude positioning relative to existing egocentric, mixed-view, and embodied-learning collections. The key contributions of our work can be summarized as follows:
\input{tabs/0_comparison}

\begin{itemize}[leftmargin=*, topsep=0pt]
\item We introduce \emph{HumanNet}, a one-million-hour human-centric video corpus spanning first-person and third-person views of fine-grained physical activities, organized by a multi-axis taxonomy over source type, viewpoint, task structure, environment, interaction style, motion category, and metadata availability.
\item We describe a full curation pipeline covering acquisition, human-centric filtering, viewpoint characterization, segmentation, deduplication, quality control, privacy review, and caption and motion annotation, turning heterogeneous web video into infrastructure for representation learning, motion-aware video modeling, and embodied pretraining.
\item We empirically validate the corpus through a controlled vision-language-action post-training study, showing that 1{,}000 hours of egocentric pretraining from \emph{HumanNet} matches or modestly surpasses 100 hours of real-robot from Magic Cobot pretraining under an identical downstream regime, and substantially closes the gap to a 20{,}000-hour real-robot baseline.
\end{itemize}

%% file: tabs/0_comparison.tex
\begin{table}[t]
\caption{Illustrative comparison between \emph{HumanNet} and representative prior corpora. The comparison highlights classic egocentric and exocentric datasets as well as recent releases.}
\label{tab:comparison}
\centering
\setlength{\tabcolsep}{3.5pt}
\resizebox{\linewidth}{!}{%
\begin{tabular}{lcccc}
\toprule
Dataset & Scale & Viewpoints & Activity Scope & Embodied Use \\
\midrule
\rowcolor{RowRed}
\multicolumn{5}{l}{\textit{Ego-Centric}} \\
EPIC-KITCHENS-100~\cite{egokitchens} & $\sim$100h & First-person & Kitchen actions & Limited \\
Ego4D~\cite{ego4d} & $\sim$3,670h & First-person & Daily activities & Indirect \\
HOI4D~\cite{hoi4d} & 2.4M RGB-D frames / $>$4k sequences & First-person & Category-level HOI & Direct \\
EgoDex~\cite{egodex} & 829h & First-person & Dexterous manipulation & Direct \\
OpenEgo~\cite{openego} & 1,107h & First-person & Dexterous manipulation & Direct \\
EgoScale~\cite{egoscale} & 20,854h & First-person & Dexterous manipulation & Direct \\
EgoVerse~\cite{egoverse} & 1,362h / 80k episodes & First-person & Human demonstrations & Direct \\
\midrule
\rowcolor{RowRed}
\multicolumn{5}{l}{\textit{Exo-Centric}} \\
ActivityNet~\cite{activitynet} & $>$648h & Third-person & Untrimmed human activities & Indirect \\
Kinetics~\cite{kinetics} & up to 650k clips & Third-person & Human actions & Indirect \\
Charades~\cite{charades} & 9,848 videos / 68.8h & Third-person & Indoor daily activities & Indirect \\
AVA~\cite{ava} & 430 clips / 107.5h & Third-person & Atomic visual actions & Indirect \\
Something-Something V2~\cite{something} & 220,847 videos & Third-person & Fine-grained interactions & Indirect \\
HACS~\cite{hacs} & 1.5M clips / 139k segments & Third-person & Human action clips & Indirect \\
FineGym~\cite{finegym} & 3 labeled scales (Gym99/288/530) & Third-person & Fine-grained gymnastics & Indirect \\
HowTo100M~\cite{howto100m} & 136M clips / 1.22M videos & Mostly third-person & Instructional procedures & Indirect \\
Ego-Exo4D~\cite{egoexo4d} & 1,286h & First + third & Skilled activities & Indirect \\
Human2Robot (H\&R)~\cite{human2robot} & 2,600 episodes & Third-person & Robot-action learning from human demos & Direct \\
\midrule
\textbf{Ours} & 1,000,000h & First + third & Fine-grained human activity & Direct \\
\bottomrule
\end{tabular}%
}
\end{table}

%% file: sec/2_related_work.tex
\section{Related Work}

\textbf{Human-centric activity datasets.}
Human activity data have long provided a foundation for learning visual, temporal, and physical structure from naturally occurring behavior. Third-person datasets such as ActivityNet~\cite{activitynet}, Kinetics~\cite{kinetics}, Charades~\cite{charades}, AVA~\cite{ava}, and Something-Something~\cite{something} cover broad actions, household activities, localized human behavior, and object-centric temporal reasoning. First-person datasets such as EPIC-KITCHENS~\cite{egokitchens} and Ego4D~\cite{ego4d} expose actor-centered intent, hand-object contact, and long-form everyday procedures, while Ego-Exo4D~\cite{egoexo4d} and Assembly101~\cite{assembly101} show the value of combining egocentric and exocentric viewpoints for skilled activity understanding. Dense interaction datasets such as HOI4D~\cite{hoi4d} and DexYCB~\cite{dexycb} further emphasize hand-object geometry, pose, and category-level manipulation structure. These datasets motivate a broader human-centric view in which first-person and third-person video are complementary: the former captures execution-centered cues, while the latter captures full-body motion, scene context, and interactions among people and objects. HumanNet follows this direction but targets substantially larger scale and broader activity coverage, with metadata designed for semantic, motion-aware, and interaction-aware learning.

\textbf{Robot learning from human data.}
Human data provide a complementary source of supervision for robot learning because people naturally demonstrate diverse manipulation, tool use, locomotion, and procedural behavior at a scale that is difficult to collect directly on robots. Prior work has used passive human video and broad visual pretraining to learn representations that transfer to downstream control~\cite{r3m}. More recent efforts explicitly connect human activity traces to robot learning: EgoScale~\cite{egoscale} studies scaling egocentric human data for dexterous manipulation, EgoVerse~\cite{egoverse} builds a shared egocentric data ecosystem for robot learning, and EgoMimic~\cite{egomimic} aligns human egocentric traces with robot demonstrations for imitation learning. Open vision-language-action systems such as GR00T N1~\cite{gr00t} mix heterogeneous robot logs with human video, while the Being-H series~\cite{beingh0,beingh05,beingh07} explores human interaction traces as a substrate for cross-embodiment learning and embodied foundation models. These works support the premise that human-centric video can supply scalable priors for physical intelligence, but they also highlight the need for datasets that preserve viewpoint, hand, body, object, and motion structure rather than treating human video as generic visual data.

\input{figs/1_taxonomy}

%% file: figs/1_taxonomy.tex
\begin{figure}[t]
\centering
\includegraphics[width=\linewidth]{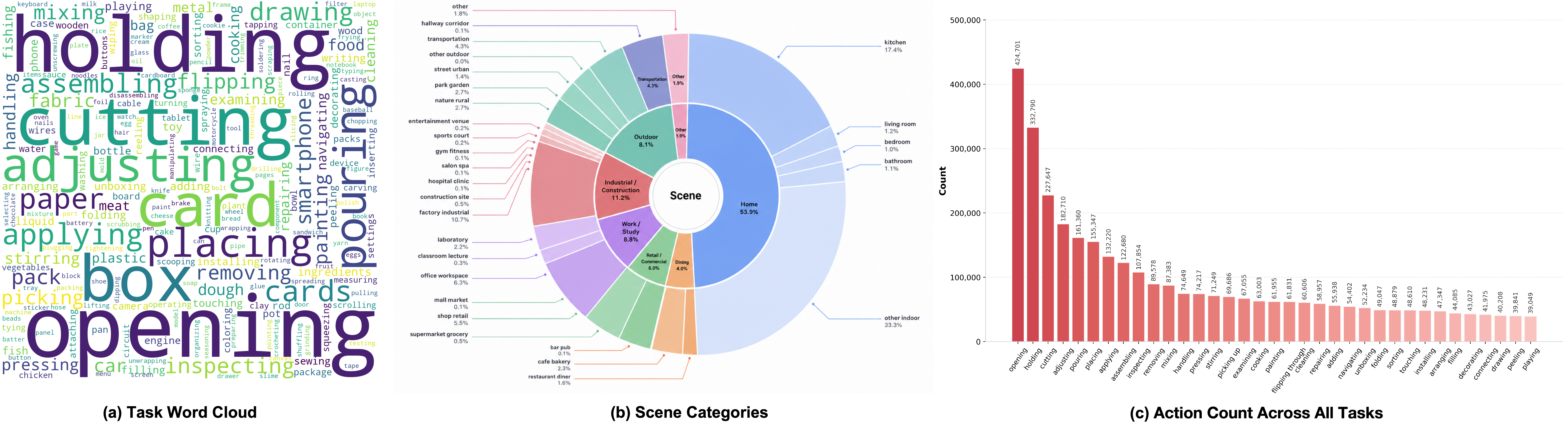}
\caption{Illustrative view of the dataset taxonomy. The corpus is organized by multiple axes rather than by a single task label or viewpoint, allowing scale to coexist with physical specificity.}
\label{fig:taxonomy}
\end{figure}

%% file: sec/3_dataset.tex
\section{The 1M-Hour Human-Centric Video Dataset}\label{sec:dataset}
Human behavior is one of the most scalable sources of data for learning physical intelligence. Humans routinely perform long-horizon interaction across diverse objects, environments, body configurations, and task variations at a scale that far exceeds what can be collected through robot teleoperation alone. \emph{HumanNet} therefore treats large-scale human-centric video as the primary data source: first-person recordings capture actor-centered intent and hand-object contact, while third-person recordings capture full-body motion, spatial context, multi-person interaction, and the geometry of activity in the surrounding scene. The dataset transforms raw heterogeneous recordings into a structured resource with caption labels, fine-grained motion annotations, hand and body signals, and motion-centric representations suitable for downstream learning.

\subsection{What Makes Human-Centric Video Suitable for Embodied Learning?}
We define \emph{human-centric video} as footage in which human activity is the organizing signal of the clip. A clip may be first-person or third-person, but it must contain physically meaningful behavior such as manipulating objects, using tools, navigating through task-relevant space, assembling or disassembling items, operating appliances or interfaces, transporting objects, coordinating with other people, or executing multi-step procedures with visible state changes in the environment. This definition intentionally excludes large volumes of passive or weakly grounded video in which human motion is incidental, the activity is not temporally coherent, or the recording lacks useful visual evidence for action, motion, or interaction.
\input{figs/2_data_pipeline}
The dataset is designed around four principles. \textbf{Scale} means that the dataset should be large enough to support long-tail coverage over activities, environments, body motions, and interaction styles, rather than saturating on a narrow task family. \textbf{Viewpoint diversity} means that first-person and third-person sources are both retained and explicitly indexed, allowing models to learn complementary actor-centered and observer-centered cues. \textbf{Physical relevance} means that the data should preserve cues useful for embodied learning, including hand-object proximity, full-body motion, state changes, action ordering, procedural structure, and scene context. \textbf{Pretraining readiness} means that the dataset must be organized so it can support modern large-scale training pipelines, including chunking, metadata indexing, quality filtering, caption labels, motion annotations, and optional alignment with text or structured labels.

At one-million-hour scale, the goal is not to claim perfect uniformity. Instead, the corpus provides the breadth needed for representations to learn invariant physical structure across heterogeneous settings and viewpoints. Compared with previous smaller embodied datasets, it covers a broader range of object frequencies, motion styles, task decompositions, social contexts, and environmental variation. Compared with generic internet video, it is more tightly aligned with human action execution, fine-grained activity semantics, and physically meaningful motion.

\subsection{Scalable Data Sources}
At the one-million-hour scale, the dataset must be heterogeneous by construction. Rather than treating this heterogeneity as noise, we index the corpus through a small set of factors that determine its value for human-centric video learning: where the data comes from, which viewpoint it uses, what kind of physical activity it contains, and what supervision signals are available after processing. Controlled and semi-structured collections provide cleaner motion and stronger metadata, while community, web-scale, and domain-specific sources expand diversity and long-tail coverage.

Interaction content is organized around physically grounded behavior rather than a closed set of semantic labels. The main emphasis is on manipulation, tool use, object transport, locomotion, full-body movement, environment state changes, multi-person coordination, and long-horizon procedures that combine motion with human-object or human-scene interaction. Many clips naturally combine several of these behaviors, so the annotation is multi-label rather than mutually exclusive.

Scene context is retained because environments change object priors, action affordances, clutter statistics, occlusions, camera motion patterns, and the visibility of body parts. Metadata is tracked separately: some sources include narrations, timestamps, or task descriptions, while others are enriched through pseudo-labels such as hand tracks, body pose, motion categories, contact estimates, scene tags, caption labels, or procedural boundaries. This structure supports flexible training mixtures without forcing all sources into a single annotation regime.

\subsection{Data Pipeline}
Figure~\ref{fig:pipeline} summarizes the end-to-end construction pipeline, which is organized into three stages: \emph{data collection}, \emph{data processing}, and \emph{annotation}. This staged design cleanly separates source acquisition from clip-level cleaning and from supervision generation, so that each stage can be audited, extended, or rerun independently as the corpus scales toward one-million-hour coverage.

\textbf{Data collection.} The collection stage couples keyword discovery with content search and retrieval. A small set of seed keywords is iteratively enlarged through keyword expansion, keyword-based crawling and cleaning, channel-level crawling, and integration of existing data sources, producing a unified keyword repository that drives subsequent retrieval. Guided by this repository, the pipeline gathers candidates from video-platform search, general web search engines, directly crawled videos, open-source datasets, and self-collection under real-world environments, which are merged into a single pool of mixed videos. The self-collected stream complements web-scale acquisition by capturing controlled first- and third-person recordings in everyday settings, providing tighter coverage of underrepresented activities, viewpoints, and scenes that are difficult to source reliably from public platforms. At this stage, channel-level and source-level filtering removes off-topic, low-quality, or passively observational sources; duplicate source entries and obviously unusable recordings are also pruned before downstream processing. For first-person material this yields an ego-video URL pool, while third-person material is retained when human motion and activity remain visually central.
\input{figs/3_samples}

\textbf{Data processing.} The processing stage converts raw videos into clip-level training samples and applies all quality control needed for downstream use. Each video is passed through de-duplication and normalization to remove near-identical copies and to unify frame rate, resolution, and container format; content filtering to retain clips with meaningful human action and observable motion; quality filtering to discard recordings with severe motion blur, heavy occlusion, static framing, or other defects that undermine learning; scene splitting that segments long videos at visual changes so that unrelated activities are not merged into a single sample; and finally video clipping that produces fixed-granularity segments. Together, these steps replace the original heterogeneous recordings with a clean, well-bounded population of clips suitable for annotation.

\textbf{Annotation.} The annotation stage enriches the processed clips with both geometric and semantic supervision. 3D hand and body pose detection recovers fine-grained motion structure; monocular SLAM estimates camera trajectory for first-person clips that satisfy stability and parallax requirements; and a retargeting module aligns recovered human motion with a unified humanoid skeleton, designating clips as robot-ready when the retargeting error remains below 15\,mm and valid-frame coverage exceeds 60\%. In parallel, an LLM-assisted captioning module produces video captions, motion descriptions, and activity classifications, which are normalized against any narrations or metadata inherited from the source. These annotations connect pixels to motion geometry, robot-relevant kinematics, and activity semantics, rather than treating the videos as unlabeled visual streams.

The pipeline therefore yields a large-scale human-centric dataset with diverse scenes, caption labels, motion annotations, hand and body metadata, and robot-ready subsets where reliable retargeting signals are available; representative clips drawn from the resulting corpus are shown in Figure~\ref{fig:samples}. Corpus-level statistics summarize the number of videos, total duration, scene count, annotated hand or pose frames, retargetable segments, and environment diversity. Privacy-sensitive content, unsafe material, and license constraints are reviewed within the same release pipeline, since both first-person and third-person recordings can contain identifiable people, private spaces, documents, screens, or proprietary workflows.
\input{figs/4_stats}
\subsection{Statistical Analysis}
We summarize the one-million-hour corpus along two complementary axes. Figure~\ref{fig:taxonomy} characterizes its \emph{semantic coverage}, that is, the activities, objects, and scenes the data spans, while Figure~\ref{fig:stats} characterizes its \emph{distributional structure}, that is, how individual clips behave under the processed pose and motion signals. Read together, the two views show a corpus that is broad along semantic axes and stratified along physical-quality axes.

Figure~\ref{fig:taxonomy} reports the lexical, scene, and category-level composition of the corpus. The action vocabulary is dominated by physically grounded manipulation verbs acting on recurring everyday objects, consistent with the design intent that the corpus emphasizes contact-rich, transformation-inducing behavior rather than passive observation. The scene hierarchy spreads clips across a wide range of indoor and outdoor environments instead of concentrating on a single domain, and the activity-category distribution exhibits a pronounced long tail. This long-tail shape motivates the one-million-hour scale, since rare but physically informative behaviors, such as folding deformable objects, handling reflective containers, or operating unfamiliar appliances, appear often enough to contribute to representation learning, whereas at smaller scales they would be easily underrepresented.

Figure~\ref{fig:stats} shifts the focus from what the corpus contains to how each clip is structured. The pose-score distribution concentrates at the high-confidence end after quality filtering, indicating that the retained clips are well suited for dense pose, hand, and motion supervision. The motion-score and motion-length distributions are both heavy-tailed yet well bounded by their statistics, reflecting a corpus dominated by short, focused interaction units while still retaining longer and more vigorous segments needed for temporal context and procedural learning. The per-category breakdown further makes the heterogeneity of the corpus explicit, with athletic and outdoor families showing longer, higher-magnitude motion, while daily activities and game-character actions concentrate on shorter, finer-grained segments.

Read jointly, these statistics expose a corpus that is broad along semantic axes and heterogeneous along physical-quality axes. High-confidence, well-segmented subsets concentrate the supervision needed for grounding, whereas the heavier-tailed regions supply the scale needed for long-tail behaviors. Exposing this structure enables mixed-supervision training recipes that match each downstream task to the appropriate slice of the corpus, a property we exploit in the downstream experiments that follow.

\subsection{Validation of Egocentric Data}
\input{figs/loss}

To test whether egocentric human video provides a transferable initialization for embodied policy learning, we conduct a controlled post-training comparison under the same LingBot-VLA architecture~\cite{lingbotvla}. The comparison isolates the effect of the pretraining source while keeping the policy architecture and the downstream data fixed. We evaluate four configurations: a Qwen-based VLM, the same Qwen VLM adapted with 100 hours of real-robot CoBot data, a Qwen VLM adapted with 1{,}000 hours of egocentric human video, and LingBot, whose Qwen backbone is trained with 20{,}000 hours of real-robot data. All variants are post-trained on the same downstream corpus of 100 tasks with 20 episodes per task, totaling 34 hours of robot interaction data. The post-training protocol follows the LingBot-VLA design but differs in how the pretrained components are initialized: for LingBot, we directly use its pretrained VLM and action expert; for the other three configurations, we use the corresponding fine-tuned VLM together with a reinitialized action expert.

Figure~\ref{fig:egocentric_validation_loss} reports validation loss across five held-out task groups under this fixed-data setting. Two observations emerge. First, the egocentric-pretrained variant consistently narrows the gap between generic web-scale language-vision initialization and robot-specialized initialization, indicating that first-person human video captures actor-centered cues, hand-object contact patterns, and procedural structure that remain useful after transfer to robot post-training. Second, although it never observes a real robot during pretraining, the model initialized with 1{,}000 hours of egocentric video matches and on several task groups slightly surpasses the model initialized with 100 hours of real-robot CoBot data, suggesting that egocentric human video is a more scalable and cost-effective substitute when teleoperated robot data is limited. Together, these results support the central design choice of \emph{HumanNet}: large-scale egocentric data is not merely an additional source of visual diversity, but a scalable bridge between internet-scale perception and embodied action learning.

%% file: figs/2_data_pipeline.tex
\begin{figure}[t]
\centering
\includegraphics[width=\linewidth]{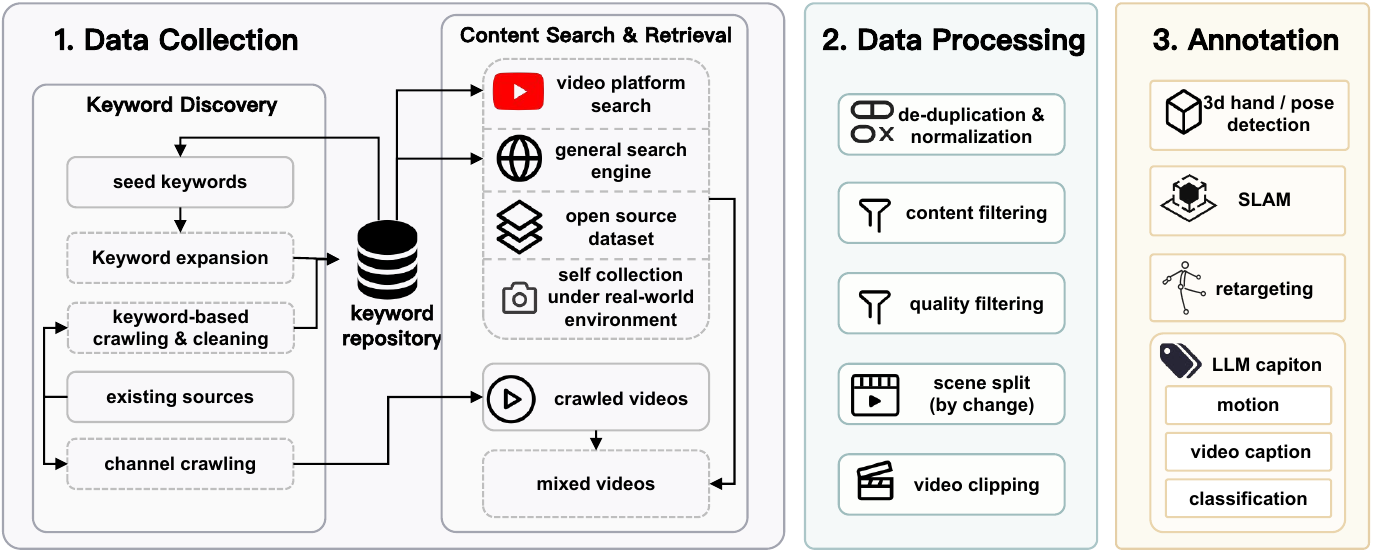}
\caption{Overview of the HumanNet data pipeline. The pipeline is organized into three stages. \textbf{(1) Data Collection} couples keyword discovery, including seed keywords, keyword expansion, keyword-based crawling, channel crawling, and existing sources, with content search and retrieval over video platforms, general search engines, open-source datasets, and self-collection under real-world environments, yielding a unified pool of mixed videos. \textbf{(2) Data Processing} converts raw videos into clip-level samples through de-duplication and normalization, content filtering, quality filtering, scene splitting by visual change, and video clipping. \textbf{(3) Annotation} enriches the processed clips with 3D hand and body pose detection, monocular SLAM, motion retargeting, and LLM-assisted captioning that produces video captions, motion descriptions, and activity classifications, resulting in a large-scale human-centric dataset with diverse scenes and robot-ready subsets.}
\label{fig:pipeline}
\end{figure}

%% file: figs/3_samples.tex
\begin{figure}[t]
\centering
\includegraphics[width=\linewidth]{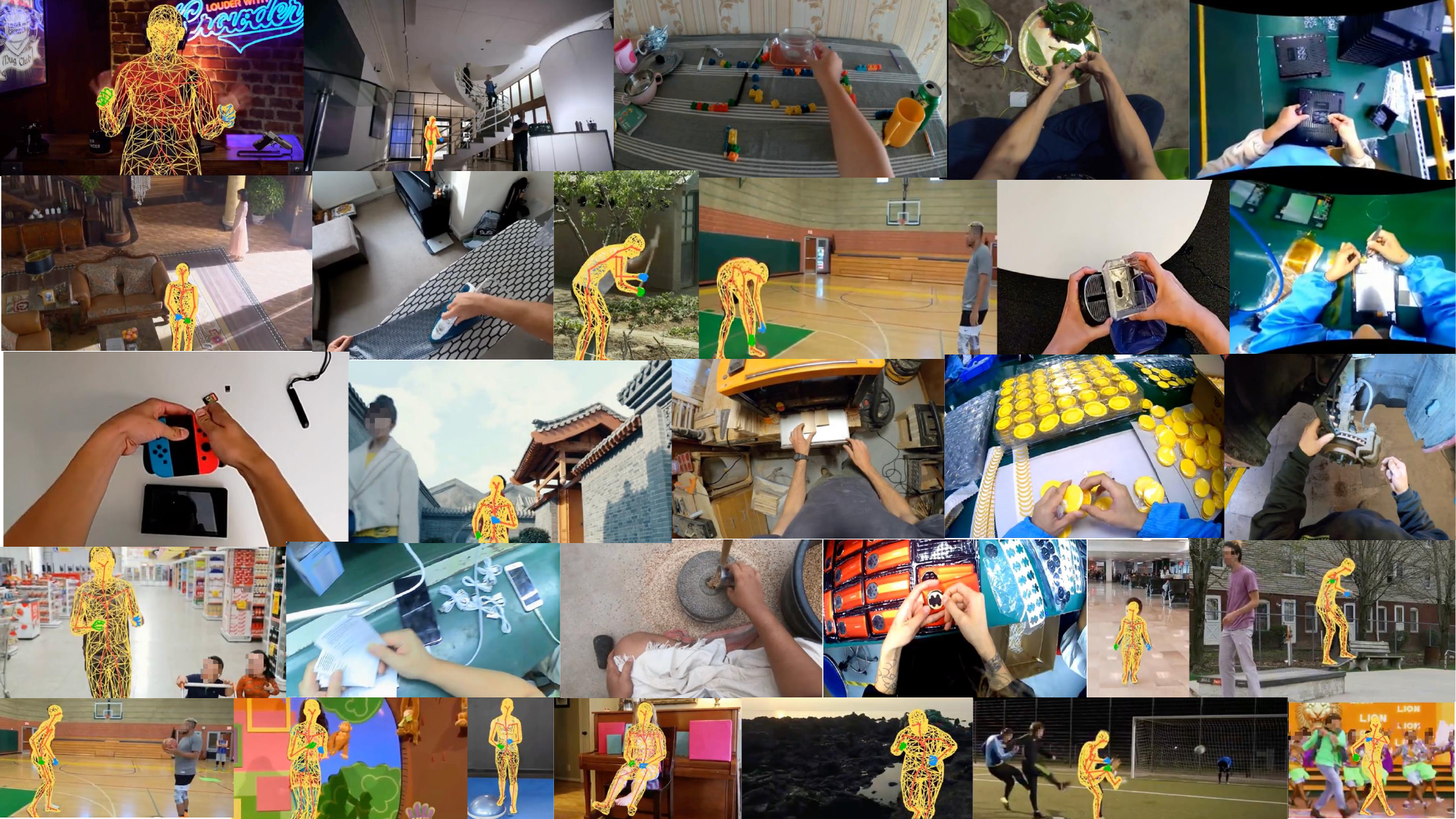}
\caption{Illustrative samples from the one-million-hour corpus. The final figure will show a diverse montage of first-person and third-person segments illustrating manipulation, tool use, locomotion, full-body motion, social interaction, and procedural tasks.}
\label{fig:samples}
\end{figure}

%% file: figs/4_stats.tex
\begin{figure}[t]
\centering
\includegraphics[width=\linewidth]{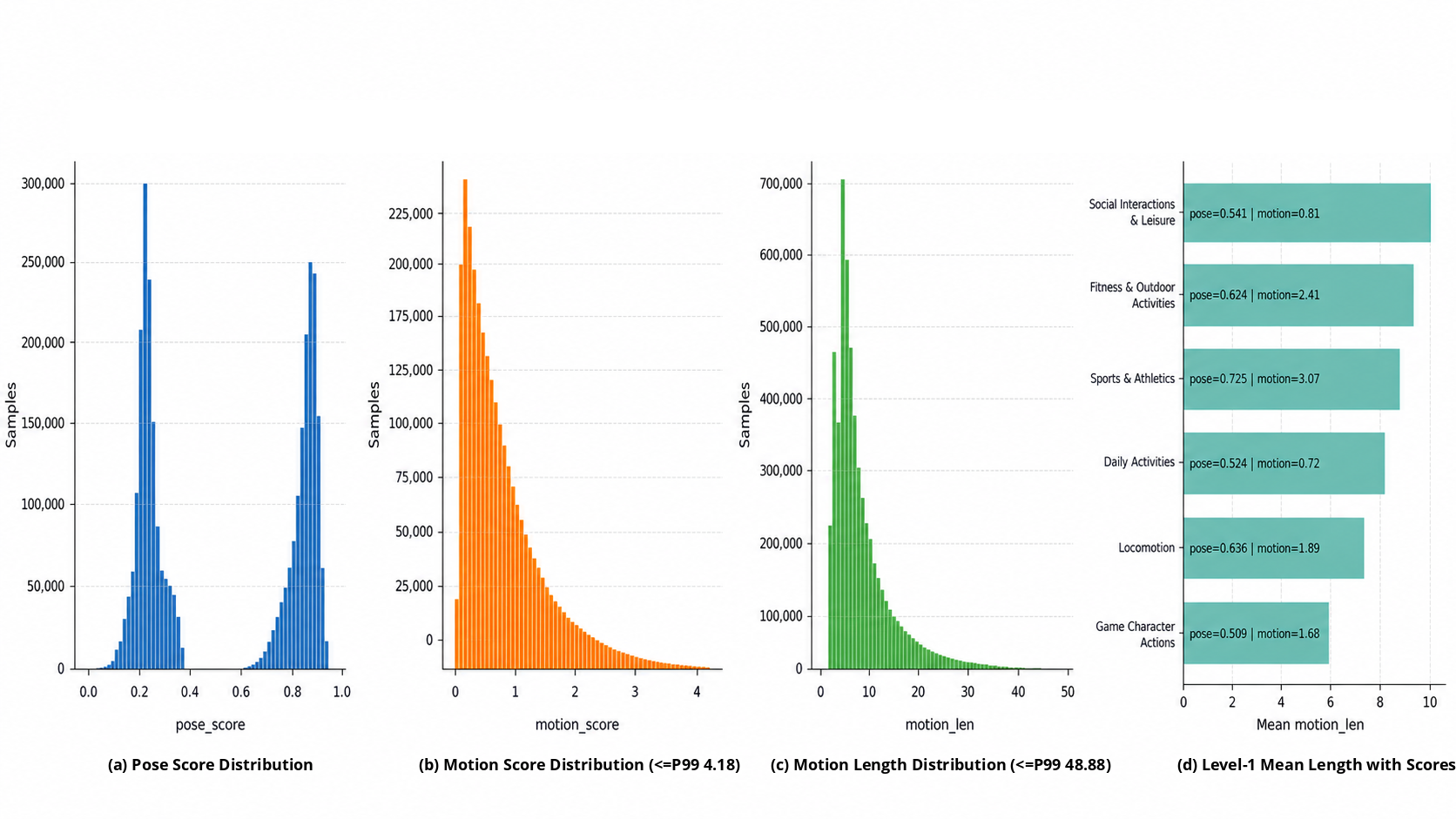}
\caption{Corpus composition statistics. The figure summarizes how the one-million-hour video corpus is distributed across source types, viewpoints, environments, activity and task categories, motion patterns, and long-tail interaction frequencies rather than only reporting a single aggregate duration.}
\label{fig:stats}
\end{figure}

%% file: figs/loss.tex
\begin{figure}[t]
\centering
\includegraphics[width=\linewidth]{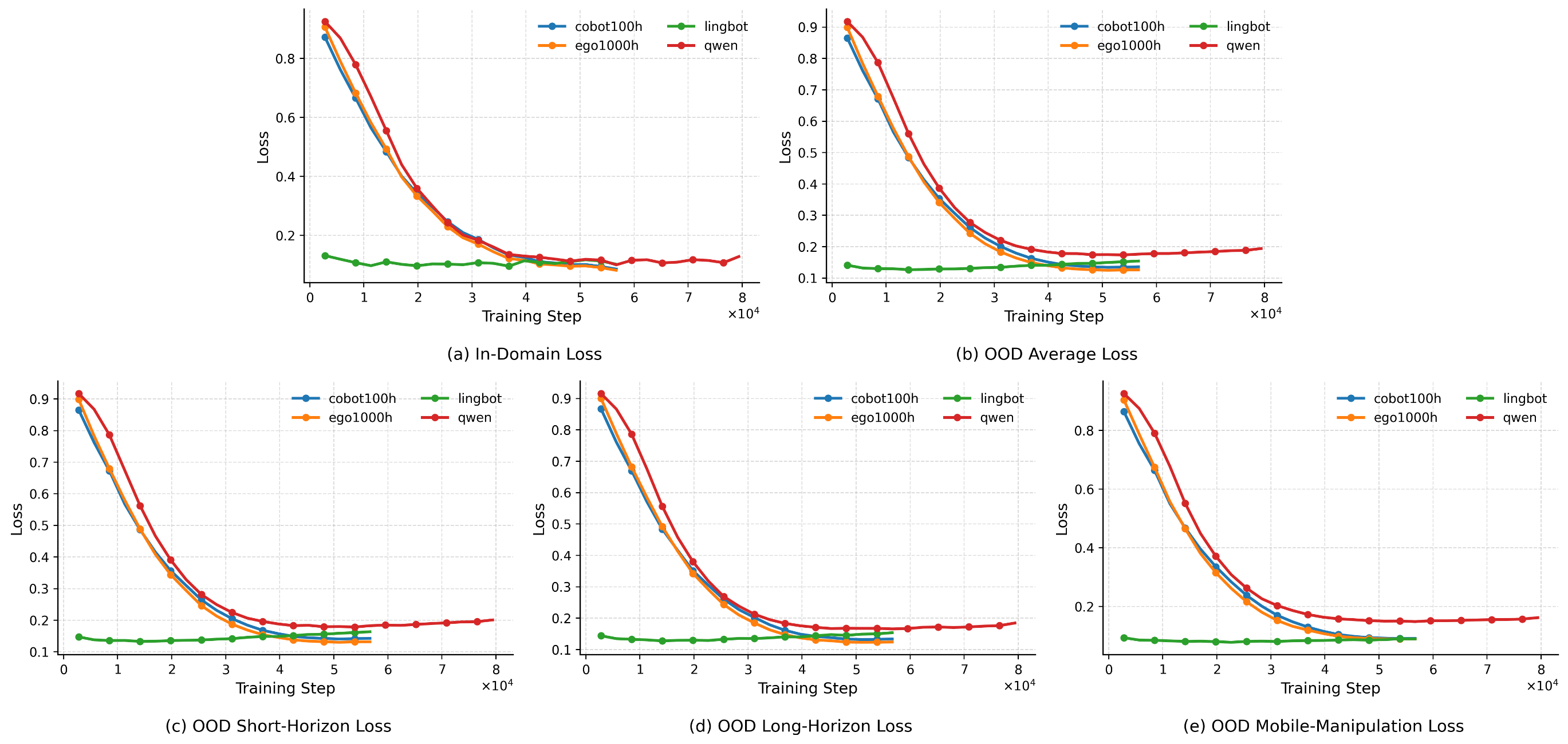}
\caption{Validation loss during controlled LingBot-VLA post-training across five held-out task groups. All configurations use the same architecture and the same 34-hour post-training corpus spanning 100 tasks with 20 episodes per task. The comparison varies only the initialization source: Qwen, Qwen adapted with 100 hours of real-robot CoBot data, Qwen adapted with 1{,}000 hours of egocentric human video, and LingBot with 20{,}000 hours of real-robot training. Lower validation loss for the egocentric-pretrained variant indicates that first-person human video provides transferable action-centric visual representations for downstream robot learning.}
\label{fig:egocentric_validation_loss}
\end{figure}

%% file: sec/4_downstream.tex
\section{Downstream Relevance}
The dataset is meant to support multiple downstream uses without committing the paper to a single benchmark suite.

\textbf{Video and VLM pretraining.}
The corpus can pretrain video encoders and video-language models that need stronger human activity, contact, and motion structure than generic internet video. First-person clips expose how actors engage objects, while third-person clips expose body pose, spatial context, and interactions among people and scenes.

\textbf{World-action model training.}
The corpus is well-suited for training world-action models that jointly capture environment dynamics and the actions that drive them. First-person clips couple actor-centered observations with hand-object contact and tool use, while third-person clips expose body motion and the resulting scene-level state changes; together with caption labels and motion annotations, this supports learning action-conditioned forward dynamics, predicting future visual states from past observations and inferred actions, and grounding language in physically executable behavior.

\textbf{Motion-aware representation learning.}
Third-person video is especially valuable for full-body motion, locomotion, posture, and multi-person dynamics, while first-person video is especially valuable for hands, contact, and actor-centered intent. Combining both viewpoints supports representations that align appearance, language, and motion rather than treating video as a sequence of independent frames.

\textbf{Human-to-robot transfer.}
This paper does not report new transfer experiments, but prior work indicates that large human datasets can supply priors when paired with alignment or action abstractions. The corpus is intended to widen the human side of that pipeline in both scale and scene diversity, while preserving motion and interaction signals that can be mapped to robot-relevant state and action representations.

\textbf{Multimodal objectives for physical AI.}
Where metadata permits, the data can support masked or predictive video modeling, language-video alignment, procedural boundary prediction, weakly supervised hand-object learning, pose and motion prediction, and caption-conditioned activity modeling. The common requirement is scale paired with annotations that preserve physically meaningful interaction structure.

%% file: sec/5_conclusion.tex
\section{Conclusion}
We present \emph{\emph{HumanNet}}, a one-million-hour human-centric video corpus that pairs first-person and third-person footage with caption labels, motion annotations, and hand and body signals, organized by a multi-axis taxonomy and produced by a curation pipeline that treats filtering, viewpoint characterization, quality control, and privacy review as first-class design choices. Under a controlled vision-language-action post-training protocol, initializing from 1{,}000 hours of egocentric video drawn from \emph{HumanNet} matches or modestly surpasses initializing from 100 hours of real-robot data and substantially closes the gap to a 20{,}000-hour real-robot baseline, indicating that egocentric human video is a scalable and cost-effective substitute when robot data is limited. Scaling diverse human activity video, with the same attention to curation and governance as to hour count, is a necessary step toward general-purpose embodied foundation models.

%% file: sec/6_limitation.tex
\section{Limitations, Ethics, and Broader Impact}
The dataset has several limitations. First, human behavior is not robot behavior. Even at one-million-hour scale, a human-centered corpus does not eliminate the embodiment gap between human hands, bodies, tools, mobility, and robot control spaces. The expected value of the dataset lies in representation learning and transferable priors, not in direct one-to-one replacement of robot data.

Second, scale introduces noise. Open-world human-centric video inevitably contains ambiguous labels, inconsistent task boundaries, missing metadata, viewpoint imbalance, and variable visual quality. Caption labels, pose estimates, and motion annotations help with coverage but introduce their own errors. This makes transparent reporting of annotation confidence and subset quality important.

Third, coverage is still uneven. A dataset can be very large while remaining biased toward certain geographies, socioeconomic contexts, occupations, camera viewpoints, body types, household routines, or public activities. Without careful analysis, one-million-hour scale can create an illusion of universality where significant blind spots remain.

Fourth, human-centric video raises serious privacy and safety issues. First-person recordings may capture bystanders, sensitive interiors, private documents, screens, or proprietary workflows. Third-person recordings may capture identifiable people, homes, workplaces, social interactions, or activities that were not originally intended for machine-learning reuse. Any public release strategy must include license review, redaction policy, restricted-content filtering, access controls where necessary, and clear documentation of what is included or excluded.

The broader impact of the dataset is dual-use. On the positive side, large-scale human-centric data may accelerate assistive systems, robotic manipulation, procedural understanding, motion modeling, and general physical AI research. On the negative side, the same data may strengthen surveillance-adjacent perception systems or enable models that inherit social and geographic biases from the source material.

%% file: main.bib
@misc{deepseekv3,
  title         = {{DeepSeek-V3} Technical Report},
  author        = {{DeepSeek-AI} and others},
  year          = {2024},
  eprint        = {2412.19437},
  archivePrefix = {arXiv},
  primaryClass  = {cs.CL},
}

@misc{qwen3,
  title         = {{Qwen3} Technical Report},
  author        = {Yang, An and Li, Anfeng and Yang, Baosong and Zhang, Beichen and Hui, Binyuan and Zheng, Bo and Yu, Bowen and Gao, Chang and others},
  year          = {2025},
  eprint        = {2505.09388},
  archivePrefix = {arXiv},
  primaryClass  = {cs.CL},
}

@misc{qwen25vl,
  title         = {{Qwen2.5-VL} Technical Report},
  author        = {Bai, Shuai and Chen, Keqin and Liu, Xuejing and Wang, Jialin and Ge, Wenbin and Song, Sibo and Dang, Kai and Wang, Peng and Wang, Shijie and Tang, Jun and Zhong, Humen and Zhu, Yuanzhi and Yang, Mingkun and Li, Zhaohai and Wan, Jianqiang and Wang, Pengfei and Ding, Wei and Fu, Zheren and Xu, Yiheng and Ye, Jiabo and Zhang, Xi and Xie, Tianbao and Cheng, Zesen and Zhang, Hang and Yang, Zhibo and Xu, Haiyang and Lin, Junyang},
  year          = {2025},
  eprint        = {2502.13923},
  archivePrefix = {arXiv},
  primaryClass  = {cs.CV},
}

@misc{internvl25,
  title         = {Expanding Performance Boundaries of Open-Source Multimodal Models with Model, Data, and Test-Time Scaling},
  author        = {Chen, Zhe and Wang, Weiyun and Cao, Yue and Liu, Yangzhou and Gao, Zhangwei and Cui, Erfei and Zhu, Jinguo and Ye, Shenglong and Tian, Hao and Liu, Zhaoyang and others},
  year          = {2024},
  eprint        = {2412.05271},
  archivePrefix = {arXiv},
  primaryClass  = {cs.CV},
}

@misc{gemma3,
  title         = {{Gemma 3} Technical Report},
  author        = {{Gemma Team}},
  year          = {2025},
  eprint        = {2503.19786},
  archivePrefix = {arXiv},
  primaryClass  = {cs.CL},
}

@misc{phi4multimodal,
      title={Phi-4-Mini Technical Report: Compact yet Powerful Multimodal Language Models via Mixture-of-LoRAs}, 
      author={Microsoft and : and Abdelrahman Abouelenin and Atabak Ashfaq and Adam Atkinson and Hany Awadalla and Nguyen Bach and Jianmin Bao and Alon Benhaim and Martin Cai and Vishrav Chaudhary and Congcong Chen and Dong Chen and Dongdong Chen and Junkun Chen and Weizhu Chen and Yen-Chun Chen and Yi-ling Chen and Qi Dai and Xiyang Dai and Ruchao Fan and Mei Gao and Min Gao and Amit Garg and Abhishek Goswami and Junheng Hao and Amr Hendy and Yuxuan Hu and Xin Jin and Mahmoud Khademi and Dongwoo Kim and Young Jin Kim and Gina Lee and Jinyu Li and Yunsheng Li and Chen Liang and Xihui Lin and Zeqi Lin and Mengchen Liu and Yang Liu and Gilsinia Lopez and Chong Luo and Piyush Madan and Vadim Mazalov and Arindam Mitra and Ali Mousavi and Anh Nguyen and Jing Pan and Daniel Perez-Becker and Jacob Platin and Thomas Portet and Kai Qiu and Bo Ren and Liliang Ren and Sambuddha Roy and Ning Shang and Yelong Shen and Saksham Singhal and Subhojit Som and Xia Song and Tetyana Sych and Praneetha Vaddamanu and Shuohang Wang and Yiming Wang and Zhenghao Wang and Haibin Wu and Haoran Xu and Weijian Xu and Yifan Yang and Ziyi Yang and Donghan Yu and Ishmam Zabir and Jianwen Zhang and Li Lyna Zhang and Yunan Zhang and Xiren Zhou},
      year={2025},
      eprint={2503.01743},
      archivePrefix={arXiv},
      primaryClass={cs.CL},
      url={https://arxiv.org/abs/2503.01743}, 
}

@article{rt1,
  title={Rt-1: Robotics transformer for real-world control at scale},
  author={Brohan, Anthony and Brown, Noah and Carbajal, Justice and Chebotar, Yevgen and Dabis, Joseph and Finn, Chelsea and Gopalakrishnan, Keerthana and Hausman, Karol and Herzog, Alex and Hsu, Jasmine and others},
  journal={arXiv preprint arXiv:2212.06817},
  year={2022}
}

@inproceedings{rt2,
  title={Rt-2: Vision-language-action models transfer web knowledge to robotic control},
  author={Zitkovich, Brianna and Yu, Tianhe and Xu, Sichun and Xu, Peng and Xiao, Ted and Xia, Fei and Wu, Jialin and Wohlhart, Paul and Welker, Stefan and Wahid, Ayzaan and others},
  booktitle={Conference on Robot Learning},
  pages={2165--2183},
  year={2023},
  organization={PMLR}
}

@misc{droid,
      title={DROID: A Large-Scale In-The-Wild Robot Manipulation Dataset}, 
      author={Alexander Khazatsky and Karl Pertsch and Suraj Nair and Ashwin Balakrishna and Sudeep Dasari and Siddharth Karamcheti and Soroush Nasiriany and Mohan Kumar Srirama and Lawrence Yunliang Chen and Kirsty Ellis and Peter David Fagan and Joey Hejna and Masha Itkina and Marion Lepert and Yecheng Jason Ma and Patrick Tree Miller and Jimmy Wu and Suneel Belkhale and Shivin Dass and Huy Ha and Arhan Jain and Abraham Lee and Youngwoon Lee and Marius Memmel and Sungjae Park and Ilija Radosavovic and Kaiyuan Wang and Albert Zhan and Kevin Black and Cheng Chi and Kyle Beltran Hatch and Shan Lin and Jingpei Lu and Jean Mercat and Abdul Rehman and Pannag R Sanketi and Archit Sharma and Cody Simpson and Quan Vuong and Homer Rich Walke and Blake Wulfe and Ted Xiao and Jonathan Heewon Yang and Arefeh Yavary and Tony Z. Zhao and Christopher Agia and Rohan Baijal and Mateo Guaman Castro and Daphne Chen and Qiuyu Chen and Trinity Chung and Jaimyn Drake and Ethan Paul Foster and Jensen Gao and Vitor Guizilini and David Antonio Herrera and Minho Heo and Kyle Hsu and Jiaheng Hu and Muhammad Zubair Irshad and Donovon Jackson and Charlotte Le and Yunshuang Li and Kevin Lin and Roy Lin and Zehan Ma and Abhiram Maddukuri and Suvir Mirchandani and Daniel Morton and Tony Nguyen and Abigail O'Neill and Rosario Scalise and Derick Seale and Victor Son and Stephen Tian and Emi Tran and Andrew E. Wang and Yilin Wu and Annie Xie and Jingyun Yang and Patrick Yin and Yunchu Zhang and Osbert Bastani and Glen Berseth and Jeannette Bohg and Ken Goldberg and Abhinav Gupta and Abhishek Gupta and Dinesh Jayaraman and Joseph J Lim and Jitendra Malik and Roberto Martín-Martín and Subramanian Ramamoorthy and Dorsa Sadigh and Shuran Song and Jiajun Wu and Michael C. Yip and Yuke Zhu and Thomas Kollar and Sergey Levine and Chelsea Finn},
      year={2025},
      eprint={2403.12945},
      archivePrefix={arXiv},
      primaryClass={cs.RO},
      url={https://arxiv.org/abs/2403.12945}, 
}

@inproceedings{activitynet,
  title={Activitynet: A large-scale video benchmark for human activity understanding},
  author={Caba Heilbron, Fabian and Escorcia, Victor and Ghanem, Bernard and Carlos Niebles, Juan},
  booktitle={Proceedings of the ieee conference on computer vision and pattern recognition},
  pages={961--970},
  year={2015}
}

@misc{kinetics,
      title={The Kinetics Human Action Video Dataset}, 
      author={Will Kay and Joao Carreira and Karen Simonyan and Brian Zhang and Chloe Hillier and Sudheendra Vijayanarasimhan and Fabio Viola and Tim Green and Trevor Back and Paul Natsev and Mustafa Suleyman and Andrew Zisserman},
      year={2017},
      eprint={1705.06950},
      archivePrefix={arXiv},
      primaryClass={cs.CV},
      url={https://arxiv.org/abs/1705.06950}, 
}

@misc{charades,
      title={Hollywood in Homes: Crowdsourcing Data Collection for Activity Understanding}, 
      author={Gunnar A. Sigurdsson and Gül Varol and Xiaolong Wang and Ali Farhadi and Ivan Laptev and Abhinav Gupta},
      year={2016},
      eprint={1604.01753},
      archivePrefix={arXiv},
      primaryClass={cs.CV},
      url={https://arxiv.org/abs/1604.01753}, 
}

@misc{ava,
      title={AVA: A Video Dataset of Spatio-temporally Localized Atomic Visual Actions}, 
      author={Chunhui Gu and Chen Sun and David A. Ross and Carl Vondrick and Caroline Pantofaru and Yeqing Li and Sudheendra Vijayanarasimhan and George Toderici and Susanna Ricco and Rahul Sukthankar and Cordelia Schmid and Jitendra Malik},
      year={2018},
      eprint={1705.08421},
      archivePrefix={arXiv},
      primaryClass={cs.CV},
      url={https://arxiv.org/abs/1705.08421}, 
}

@misc{something,
      title={The "something something" video database for learning and evaluating visual common sense}, 
      author={Raghav Goyal and Samira Ebrahimi Kahou and Vincent Michalski and Joanna Materzyńska and Susanne Westphal and Heuna Kim and Valentin Haenel and Ingo Fruend and Peter Yianilos and Moritz Mueller-Freitag and Florian Hoppe and Christian Thurau and Ingo Bax and Roland Memisevic},
      year={2017},
      eprint={1706.04261},
      archivePrefix={arXiv},
      primaryClass={cs.CV},
      url={https://arxiv.org/abs/1706.04261}, 
}

@misc{howto100m,
      title={HowTo100M: Learning a Text-Video Embedding by Watching Hundred Million Narrated Video Clips}, 
      author={Antoine Miech and Dimitri Zhukov and Jean-Baptiste Alayrac and Makarand Tapaswi and Ivan Laptev and Josef Sivic},
      year={2019},
      eprint={1906.03327},
      archivePrefix={arXiv},
      primaryClass={cs.CV},
      url={https://arxiv.org/abs/1906.03327}, 
}

@misc{egodex,
      title={EgoDex: Learning Dexterous Manipulation from Large-Scale Egocentric Video}, 
      author={Ryan Hoque and Peide Huang and David J. Yoon and Mouli Sivapurapu and Jian Zhang},
      year={2026},
      eprint={2505.11709},
      archivePrefix={arXiv},
      primaryClass={cs.CV},
      url={https://arxiv.org/abs/2505.11709}, 
}

@misc{openego,
      title={OpenEgo: A Large-Scale Multimodal Egocentric Dataset for Dexterous Manipulation}, 
      author={Ahad Jawaid and Yu Xiang},
      year={2025},
      eprint={2509.05513},
      archivePrefix={arXiv},
      primaryClass={cs.CV},
      url={https://arxiv.org/abs/2509.05513}, 
}

@misc{human2robot,
      title={Human2Robot: Learning Robot Actions from Paired Human-Robot Videos}, 
      author={Sicheng Xie and Haidong Cao and Zejia Weng and Zhen Xing and Haoran Chen and Shiwei Shen and Jiaqi Leng and Zuxuan Wu and Yu-Gang Jiang},
      year={2025},
      eprint={2502.16587},
      archivePrefix={arXiv},
      primaryClass={cs.RO},
      url={https://arxiv.org/abs/2502.16587}, 
}

@misc{egoverse,
      title={EgoVerse: An Egocentric Human Dataset for Robot Learning from Around the World}, 
      author={Ryan Punamiya and Simar Kareer and Zeyi Liu and Josh Citron and Ri-Zhao Qiu and Xiongyi Cai and Alexey Gavryushin and Jiaqi Chen and Davide Liconti and Lawrence Y. Zhu and Patcharapong Aphiwetsa and Baoyu Li and Aniketh Cheluva and Pranav Kuppili and Yangcen Liu and Dhruv Patel and Aidan Gao and Hye-Young Chung and Ryan Co and Renee Zbizika and Jeff Liu and Xiaomeng Xu and Haoyu Xiong and Geng Chen and Sebastiano Oliani and Chenyu Yang and Xi Wang and James Fort and Richard Newcombe and Josh Gao and Jason Chong and Garrett Matsuda and Aseem Doriwala and Marc Pollefeys and Robert Katzschmann and Xiaolong Wang and Shuran Song and Judy Hoffman and Danfei Xu},
      year={2026},
      eprint={2604.07607},
      archivePrefix={arXiv},
      primaryClass={cs.RO},
      url={https://arxiv.org/abs/2604.07607}, 
}

@misc{egoscale,
      title={EgoScale: Scaling Dexterous Manipulation with Diverse Egocentric Human Data}, 
      author={Ruijie Zheng and Dantong Niu and Yuqi Xie and Jing Wang and Mengda Xu and Yunfan Jiang and Fernando Castañeda and Fengyuan Hu and You Liang Tan and Letian Fu and Trevor Darrell and Furong Huang and Yuke Zhu and Danfei Xu and Linxi Fan},
      year={2026},
      eprint={2602.16710},
      archivePrefix={arXiv},
      primaryClass={cs.RO},
      url={https://arxiv.org/abs/2602.16710}, 
}

@misc{hoi4d,
      title={HOI4D: A 4D Egocentric Dataset for Category-Level Human-Object Interaction}, 
      author={Yunze Liu and Yun Liu and Che Jiang and Kangbo Lyu and Weikang Wan and Hao Shen and Boqiang Liang and Zhoujie Fu and He Wang and Li Yi},
      year={2024},
      eprint={2203.01577},
      archivePrefix={arXiv},
      primaryClass={cs.CV},
      url={https://arxiv.org/abs/2203.01577}, 
}

@misc{egoschema,
      title={EgoSchema: A Diagnostic Benchmark for Very Long-form Video Language Understanding}, 
      author={Karttikeya Mangalam and Raiymbek Akshulakov and Jitendra Malik},
      year={2023},
      eprint={2308.09126},
      archivePrefix={arXiv},
      primaryClass={cs.CV},
      url={https://arxiv.org/abs/2308.09126}, 
}

@misc{ego4d,
      title={Ego4D: Around the World in 3,000 Hours of Egocentric Video}, 
      author={Kristen Grauman and Andrew Westbury and Eugene Byrne and Zachary Chavis and Antonino Furnari and Rohit Girdhar and Jackson Hamburger and Hao Jiang and Miao Liu and Xingyu Liu and Miguel Martin and Tushar Nagarajan and Ilija Radosavovic and Santhosh Kumar Ramakrishnan and Fiona Ryan and Jayant Sharma and Michael Wray and Mengmeng Xu and Eric Zhongcong Xu and Chen Zhao and Siddhant Bansal and Dhruv Batra and Vincent Cartillier and Sean Crane and Tien Do and Morrie Doulaty and Akshay Erapalli and Christoph Feichtenhofer and Adriano Fragomeni and Qichen Fu and Abrham Gebreselasie and Cristina Gonzalez and James Hillis and Xuhua Huang and Yifei Huang and Wenqi Jia and Weslie Khoo and Jachym Kolar and Satwik Kottur and Anurag Kumar and Federico Landini and Chao Li and Yanghao Li and Zhenqiang Li and Karttikeya Mangalam and Raghava Modhugu and Jonathan Munro and Tullie Murrell and Takumi Nishiyasu and Will Price and Paola Ruiz Puentes and Merey Ramazanova and Leda Sari and Kiran Somasundaram and Audrey Southerland and Yusuke Sugano and Ruijie Tao and Minh Vo and Yuchen Wang and Xindi Wu and Takuma Yagi and Ziwei Zhao and Yunyi Zhu and Pablo Arbelaez and David Crandall and Dima Damen and Giovanni Maria Farinella and Christian Fuegen and Bernard Ghanem and Vamsi Krishna Ithapu and C. V. Jawahar and Hanbyul Joo and Kris Kitani and Haizhou Li and Richard Newcombe and Aude Oliva and Hyun Soo Park and James M. Rehg and Yoichi Sato and Jianbo Shi and Mike Zheng Shou and Antonio Torralba and Lorenzo Torresani and Mingfei Yan and Jitendra Malik},
      year={2022},
      eprint={2110.07058},
      archivePrefix={arXiv},
      primaryClass={cs.CV},
      url={https://arxiv.org/abs/2110.07058}, 
}

@article{egokitchens,
  title={Rescaling egocentric vision: Collection, pipeline and challenges for epic-kitchens-100},
  author={Damen, Dima and Doughty, Hazel and Farinella, Giovanni Maria and Furnari, Antonino and Kazakos, Evangelos and Ma, Jian and Moltisanti, Davide and Munro, Jonathan and Perrett, Toby and Price, Will and others},
  journal={International Journal of Computer Vision},
  volume={130},
  number={1},
  pages={33--55},
  year={2022},
  publisher={Springer}
}

@misc{hacs,
      title={HACS: Human Action Clips and Segments Dataset for Recognition and Temporal Localization}, 
      author={Hang Zhao and Antonio Torralba and Lorenzo Torresani and Zhicheng Yan},
      year={2019},
      eprint={1712.09374},
      archivePrefix={arXiv},
      primaryClass={cs.CV},
      url={https://arxiv.org/abs/1712.09374}, 
}

@misc{finegym,
      title={FineGym: A Hierarchical Video Dataset for Fine-grained Action Understanding}, 
      author={Dian Shao and Yue Zhao and Bo Dai and Dahua Lin},
      year={2020},
      eprint={2004.06704},
      archivePrefix={arXiv},
      primaryClass={cs.CV},
      url={https://arxiv.org/abs/2004.06704}, 
}

@misc{openx,
      title={Open X-Embodiment: Robotic Learning Datasets and RT-X Models}, 
      author={Embodiment Collaboration and Abby O'Neill and Abdul Rehman and Abhinav Gupta and Abhiram Maddukuri and Abhishek Gupta and Abhishek Padalkar and Abraham Lee and Acorn Pooley and Agrim Gupta and Ajay Mandlekar and Ajinkya Jain and Albert Tung and Alex Bewley and Alex Herzog and Alex Irpan and Alexander Khazatsky and Anant Rai and Anchit Gupta and Andrew Wang and Andrey Kolobov and Anikait Singh and Animesh Garg and Aniruddha Kembhavi and Annie Xie and Anthony Brohan and Antonin Raffin and Archit Sharma and Arefeh Yavary and Arhan Jain and Ashwin Balakrishna and Ayzaan Wahid and Ben Burgess-Limerick and Beomjoon Kim and Bernhard Schölkopf and Blake Wulfe and Brian Ichter and Cewu Lu and Charles Xu and Charlotte Le and Chelsea Finn and Chen Wang and Chenfeng Xu and Cheng Chi and Chenguang Huang and Christine Chan and Christopher Agia and Chuer Pan and Chuyuan Fu and Coline Devin and Danfei Xu and Daniel Morton and Danny Driess and Daphne Chen and Deepak Pathak and Dhruv Shah and Dieter Büchler and Dinesh Jayaraman and Dmitry Kalashnikov and Dorsa Sadigh and Edward Johns and Ethan Foster and Fangchen Liu and Federico Ceola and Fei Xia and Feiyu Zhao and Felipe Vieira Frujeri and Freek Stulp and Gaoyue Zhou and Gaurav S. Sukhatme and Gautam Salhotra and Ge Yan and Gilbert Feng and Giulio Schiavi and Glen Berseth and Gregory Kahn and Guangwen Yang and Guanzhi Wang and Hao Su and Hao-Shu Fang and Haochen Shi and Henghui Bao and Heni Ben Amor and Henrik I Christensen and Hiroki Furuta and Homanga Bharadhwaj and Homer Walke and Hongjie Fang and Huy Ha and Igor Mordatch and Ilija Radosavovic and Isabel Leal and Jacky Liang and Jad Abou-Chakra and Jaehyung Kim and Jaimyn Drake and Jan Peters and Jan Schneider and Jasmine Hsu and Jay Vakil and Jeannette Bohg and Jeffrey Bingham and Jeffrey Wu and Jensen Gao and Jiaheng Hu and Jiajun Wu and Jialin Wu and Jiankai Sun and Jianlan Luo and Jiayuan Gu and Jie Tan and Jihoon Oh and Jimmy Wu and Jingpei Lu and Jingyun Yang and Jitendra Malik and João Silvério and Joey Hejna and Jonathan Booher and Jonathan Tompson and Jonathan Yang and Jordi Salvador and Joseph J. Lim and Junhyek Han and Kaiyuan Wang and Kanishka Rao and Karl Pertsch and Karol Hausman and Keegan Go and Keerthana Gopalakrishnan and Ken Goldberg and Kendra Byrne and Kenneth Oslund and Kento Kawaharazuka and Kevin Black and Kevin Lin and Kevin Zhang and Kiana Ehsani and Kiran Lekkala and Kirsty Ellis and Krishan Rana and Krishnan Srinivasan and Kuan Fang and Kunal Pratap Singh and Kuo-Hao Zeng and Kyle Hatch and Kyle Hsu and Laurent Itti and Lawrence Yunliang Chen and Lerrel Pinto and Li Fei-Fei and Liam Tan and Linxi "Jim" Fan and Lionel Ott and Lisa Lee and Luca Weihs and Magnum Chen and Marion Lepert and Marius Memmel and Masayoshi Tomizuka and Masha Itkina and Mateo Guaman Castro and Max Spero and Maximilian Du and Michael Ahn and Michael C. Yip and Mingtong Zhang and Mingyu Ding and Minho Heo and Mohan Kumar Srirama and Mohit Sharma and Moo Jin Kim and Muhammad Zubair Irshad and Naoaki Kanazawa and Nicklas Hansen and Nicolas Heess and Nikhil J Joshi and Niko Suenderhauf and Ning Liu and Norman Di Palo and Nur Muhammad Mahi Shafiullah and Oier Mees and Oliver Kroemer and Osbert Bastani and Pannag R Sanketi and Patrick "Tree" Miller and Patrick Yin and Paul Wohlhart and Peng Xu and Peter David Fagan and Peter Mitrano and Pierre Sermanet and Pieter Abbeel and Priya Sundaresan and Qiuyu Chen and Quan Vuong and Rafael Rafailov and Ran Tian and Ria Doshi and Roberto Martín-Martín and Rohan Baijal and Rosario Scalise and Rose Hendrix and Roy Lin and Runjia Qian and Ruohan Zhang and Russell Mendonca and Rutav Shah and Ryan Hoque and Ryan Julian and Samuel Bustamante and Sean Kirmani and Sergey Levine and Shan Lin and Sherry Moore and Shikhar Bahl and Shivin Dass and Shubham Sonawani and Shubham Tulsiani and Shuran Song and Sichun Xu and Siddhant Haldar and Siddharth Karamcheti and Simeon Adebola and Simon Guist and Soroush Nasiriany and Stefan Schaal and Stefan Welker and Stephen Tian and Subramanian Ramamoorthy and Sudeep Dasari and Suneel Belkhale and Sungjae Park and Suraj Nair and Suvir Mirchandani and Takayuki Osa and Tanmay Gupta and Tatsuya Harada and Tatsuya Matsushima and Ted Xiao and Thomas Kollar and Tianhe Yu and Tianli Ding and Todor Davchev and Tony Z. Zhao and Travis Armstrong and Trevor Darrell and Trinity Chung and Vidhi Jain and Vikash Kumar and Vincent Vanhoucke and Vitor Guizilini and Wei Zhan and Wenxuan Zhou and Wolfram Burgard and Xi Chen and Xiangyu Chen and Xiaolong Wang and Xinghao Zhu and Xinyang Geng and Xiyuan Liu and Xu Liangwei and Xuanlin Li and Yansong Pang and Yao Lu and Yecheng Jason Ma and Yejin Kim and Yevgen Chebotar and Yifan Zhou and Yifeng Zhu and Yilin Wu and Ying Xu and Yixuan Wang and Yonatan Bisk and Yongqiang Dou and Yoonyoung Cho and Youngwoon Lee and Yuchen Cui and Yue Cao and Yueh-Hua Wu and Yujin Tang and Yuke Zhu and Yunchu Zhang and Yunfan Jiang and Yunshuang Li and Yunzhu Li and Yusuke Iwasawa and Yutaka Matsuo and Zehan Ma and Zhuo Xu and Zichen Jeff Cui and Zichen Zhang and Zipeng Fu and Zipeng Lin},
      year={2025},
      eprint={2310.08864},
      archivePrefix={arXiv},
      primaryClass={cs.RO},
      url={https://arxiv.org/abs/2310.08864}, 
}

@misc{r3m,
      title={R3M: A Universal Visual Representation for Robot Manipulation}, 
      author={Suraj Nair and Aravind Rajeswaran and Vikash Kumar and Chelsea Finn and Abhinav Gupta},
      year={2022},
      eprint={2203.12601},
      archivePrefix={arXiv},
      primaryClass={cs.RO},
      url={https://arxiv.org/abs/2203.12601}, 
}

@misc{beingh05,
      title={Being-H0.5: Scaling Human-Centric Robot Learning for Cross-Embodiment Generalization}, 
      author={Hao Luo and Ye Wang and Wanpeng Zhang and Sipeng Zheng and Ziheng Xi and Chaoyi Xu and Haiweng Xu and Haoqi Yuan and Chi Zhang and Yiqing Wang and Yicheng Feng and Zongqing Lu},
      year={2026},
      eprint={2601.12993},
      archivePrefix={arXiv},
      primaryClass={cs.RO},
      url={https://arxiv.org/abs/2601.12993}, 
}

@misc{beingh0,
      title={Being-H0: Vision-Language-Action Pretraining from Large-Scale Human Videos}, 
      author={Hao Luo and Yicheng Feng and Wanpeng Zhang and Sipeng Zheng and Ye Wang and Haoqi Yuan and Jiazheng Liu and Chaoyi Xu and Qin Jin and Zongqing Lu},
      year={2025},
      eprint={2507.15597},
      archivePrefix={arXiv},
      primaryClass={cs.CV},
      url={https://arxiv.org/abs/2507.15597}, 
}

@misc{egoexo4d,
      title={Ego-Exo4D: Understanding Skilled Human Activity from First- and Third-Person Perspectives}, 
      author={Kristen Grauman and Andrew Westbury and Lorenzo Torresani and Kris Kitani and Jitendra Malik and Triantafyllos Afouras and Kumar Ashutosh and Vijay Baiyya and Siddhant Bansal and Bikram Boote and Eugene Byrne and Zach Chavis and Joya Chen and Feng Cheng and Fu-Jen Chu and Sean Crane and Avijit Dasgupta and Jing Dong and Maria Escobar and Cristhian Forigua and Abrham Gebreselasie and Sanjay Haresh and Jing Huang and Md Mohaiminul Islam and Suyog Jain and Rawal Khirodkar and Devansh Kukreja and Kevin J Liang and Jia-Wei Liu and Sagnik Majumder and Yongsen Mao and Miguel Martin and Effrosyni Mavroudi and Tushar Nagarajan and Francesco Ragusa and Santhosh Kumar Ramakrishnan and Luigi Seminara and Arjun Somayazulu and Yale Song and Shan Su and Zihui Xue and Edward Zhang and Jinxu Zhang and Angela Castillo and Changan Chen and Xinzhu Fu and Ryosuke Furuta and Cristina Gonzalez and Prince Gupta and Jiabo Hu and Yifei Huang and Yiming Huang and Weslie Khoo and Anush Kumar and Robert Kuo and Sach Lakhavani and Miao Liu and Mi Luo and Zhengyi Luo and Brighid Meredith and Austin Miller and Oluwatumininu Oguntola and Xiaqing Pan and Penny Peng and Shraman Pramanick and Merey Ramazanova and Fiona Ryan and Wei Shan and Kiran Somasundaram and Chenan Song and Audrey Southerland and Masatoshi Tateno and Huiyu Wang and Yuchen Wang and Takuma Yagi and Mingfei Yan and Xitong Yang and Zecheng Yu and Shengxin Cindy Zha and Chen Zhao and Ziwei Zhao and Zhifan Zhu and Jeff Zhuo and Pablo Arbelaez and Gedas Bertasius and David Crandall and Dima Damen and Jakob Engel and Giovanni Maria Farinella and Antonino Furnari and Bernard Ghanem and Judy Hoffman and C. V. Jawahar and Richard Newcombe and Hyun Soo Park and James M. Rehg and Yoichi Sato and Manolis Savva and Jianbo Shi and Mike Zheng Shou and Michael Wray},
      year={2024},
      eprint={2311.18259},
      archivePrefix={arXiv},
      primaryClass={cs.CV},
      url={https://arxiv.org/abs/2311.18259}, 
}

@misc{egomimic,
      title={EgoMimic: Scaling Imitation Learning via Egocentric Video}, 
      author={Simar Kareer and Dhruv Patel and Ryan Punamiya and Pranay Mathur and Shuo Cheng and Chen Wang and Judy Hoffman and Danfei Xu},
      year={2024},
      eprint={2410.24221},
      archivePrefix={arXiv},
      primaryClass={cs.RO},
      url={https://arxiv.org/abs/2410.24221}, 
}

@article{deng2026rethinking,
  title={Rethinking Video Generation Model for the Embodied World},
  author={Deng, Yufan and Pan, Zilin and Zhang, Hongyu and Li, Xiaojie and Hu, Ruoqing and Ding, Yufei and Zou, Yiming and Zeng, Yan and Zhou, Daquan},
  journal={arXiv preprint arXiv:2601.15282},
  year={2026}
}

@misc{gr00t,
      title={GR00T N1: An Open Foundation Model for Generalist Humanoid Robots}, 
      author={NVIDIA and : and Johan Bjorck and Fernando Castañeda and Nikita Cherniadev and Xingye Da and Runyu Ding and Linxi "Jim" Fan and Yu Fang and Dieter Fox and Fengyuan Hu and Spencer Huang and Joel Jang and Zhenyu Jiang and Jan Kautz and Kaushil Kundalia and Lawrence Lao and Zhiqi Li and Zongyu Lin and Kevin Lin and Guilin Liu and Edith Llontop and Loic Magne and Ajay Mandlekar and Avnish Narayan and Soroush Nasiriany and Scott Reed and You Liang Tan and Guanzhi Wang and Zu Wang and Jing Wang and Qi Wang and Jiannan Xiang and Yuqi Xie and Yinzhen Xu and Zhenjia Xu and Seonghyeon Ye and Zhiding Yu and Ao Zhang and Hao Zhang and Yizhou Zhao and Ruijie Zheng and Yuke Zhu},
      year={2025},
      eprint={2503.14734},
      archivePrefix={arXiv},
      primaryClass={cs.RO},
      url={https://arxiv.org/abs/2503.14734}, 
}

@misc{rh20t,
      title={RH20T: A Comprehensive Robotic Dataset for Learning Diverse Skills in One-Shot}, 
      author={Hao-Shu Fang and Hongjie Fang and Zhenyu Tang and Jirong Liu and Chenxi Wang and Junbo Wang and Haoyi Zhu and Cewu Lu},
      year={2023},
      eprint={2307.00595},
      archivePrefix={arXiv},
      primaryClass={cs.RO},
      url={https://arxiv.org/abs/2307.00595}, 
}

@misc{beingh07,
      title={Being-H0.7: A Latent World-Action Model from Egocentric Videos}, 
      author={Hao Luo and Wanpeng Zhang and Yicheng Feng and Sipeng Zheng and Haiweng Xu and Chaoyi Xu and Ziheng Xi and Yuhui Fu and Zongqing Lu},
      year={2026},
      eprint={2605.00078},
      archivePrefix={arXiv},
      primaryClass={cs.RO},
      url={https://arxiv.org/abs/2605.00078}, 
}

@inproceedings{assembly101,
  title={Assembly101: A large-scale multi-view video dataset for understanding procedural activities},
  author={Sener, Fadime and Chatterjee, Dibyadip and Shelepov, Daniel and He, Kun and Singhania, Dipika and Wang, Robert and Yao, Angela},
  booktitle={Proceedings of the IEEE/CVF Conference on Computer Vision and Pattern Recognition},
  pages={21096--21106},
  year={2022}
}

@inproceedings{dexycb,
  title={Dexycb: A benchmark for capturing hand grasping of objects},
  author={Chao, Yu-Wei and Yang, Wei and Xiang, Yu and Molchanov, Pavlo and Handa, Ankur and Tremblay, Jonathan and Narang, Yashraj S and Van Wyk, Karl and Iqbal, Umar and Birchfield, Stan and others},
  booktitle={Proceedings of the IEEE/CVF conference on computer vision and pattern recognition},
  pages={9044--9053},
  year={2021}
}

@article{lingbotvla,
  title={A Pragmatic VLA Foundation Model},
  author={Wu, Wei and Lu, Fan and Wang, Yunnan and Yang, Shuai and Liu, Shi and Wang, Fangjing and Zhu, Qian and Sun, He and Wang, Yong and Ma, Shuailei and others},
  journal={arXiv preprint arXiv:2601.18692},
  year={2026}
}
